\documentclass[runningheads]{llncs}
\usepackage[T1]{fontenc}
\usepackage{graphicx}
\usepackage{url}
\usepackage{booktabs}
\usepackage{xcolor}
\usepackage{amsmath} 
\usepackage{amssymb}
\begin{document}
\title{Error- and Prediction-Driven Motor Learning in the Cortico-Cerebellar Loop}

\author{Ana Carolina Filipe\inst{1} \and Rui Ponte Costa\inst{2} \and Cláudia Soares\inst{1}}

\institute{
Department of Computer Science, NOVA School of Science
and Technology, Caparica, Portugal
\email{ac.filipe@campus.fct.unl.pt}
\and
Centre for Neural Circuits and Behaviour, University of Oxford
}

\maketitle              
\begin{abstract}

Robust control under delayed sensory feedback remains a key challenge in both robotics and neuroscience. 
Classical cerebellar models explain delay compensation through forward prediction but fail to account for fast online corrections and rapid adaptation observed in biological systems. 

We propose a cerebellum-inspired control framework that combines multiplexed predictive representations with internal feedback. By jointly encoding kinematic variables and task-relevant error signals, the model enables accurate online correction despite delayed feedback. Furthermore, incorporating feedback within the cerebellar loop significantly accelerates adaptation, reducing learning time by an order of magnitude.

Our results show that single-signal predictions are insufficient under delay, while multiplexing and feedback together provide a unified mechanism for online control and rapid learning.

\end{abstract}
\section{Introduction}

Robust feedback control is essential for intelligent systems operating in dynamic environments. A central challenge in such settings is coping with sensory delays, which can degrade performance and hinder both online correction and learning. Effective controllers must therefore balance predictive mechanisms with feedback-driven adaptation.

Biological motor systems excel at this problem. Despite significant delays in sensory pathways, they achieve both rapid error correction within a movement and fast adaptation across trials~\cite{Scott2004,Shadmehr2010}. This capability is widely attributed to the interaction between motor cortex and cerebellum. The cerebellum, in particular, is thought to implement internal forward models that predict the sensory consequences of motor commands, enabling compensation for delayed feedback~\cite{Ito2005,Kawato1995,Raymond2018,Wolpert1998}.

Classical theories, such as the Marr–Albus–Ito framework, formalize this predictive role through error-driven learning mechanisms~\cite{Marr1969,albus1971theory,Ito1984}. However, these models alone do not fully explain how biological systems achieve both rapid online corrections and stable, trial-by-trial adaptation. Recent experimental findings suggest a more complex picture: cerebellar outputs multiplex multiple task-relevant signals, including kinematics and performance errors~\cite{popa2019cerebellum,Herzfeld2015}, while motor cortex exhibits distinct feedforward and feedback-related dynamics during adaptation~\cite{perich2018rapid}. At the same time, recurrent neural network (RNN) models can reproduce aspects of motor behavior, but typically rely on direct sensory feedback to cortical units, bypassing cerebellar processing~\cite{sussillo2015neural,feulner2025neural}.

This raises a key question: \emph{how should delayed sensory feedback be routed and represented within cortico-cerebellar circuits to support both stable control and rapid adaptation?}

\begin{figure}[h!]
\centering
\includegraphics[width=0.98\textwidth]{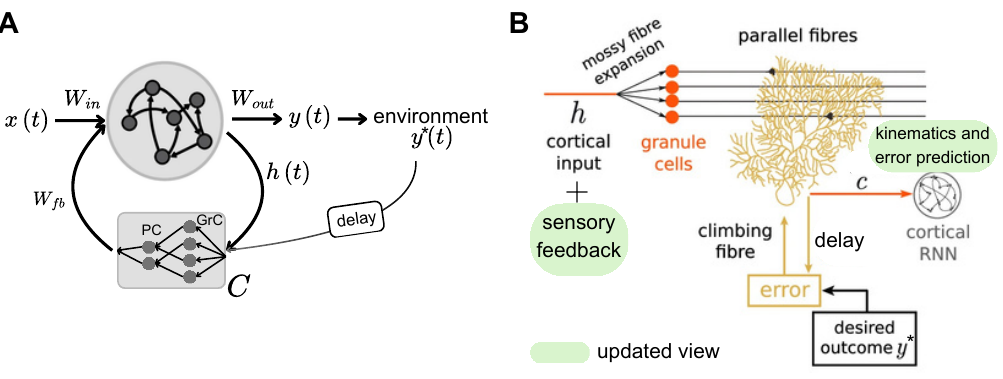}
\caption{\textbf{Model architecture and learning mechanism.} 
(A) Schematic of the proposed cortico-cerebellar model. 
(B) The cerebellar module learns through temporally specific, error-driven plasticity consistent with experimental observations. In particular, the discrepancy between the cerebellar prediction $c$ and delayed behavioral variables (120 ms) drives synaptic updates via climbing fiber signals at the parallel fibre–Purkinje cell synapses.}
\label{fig:model}
\end{figure}

We address this question with a recurrent neural network model augmented by a cerebellum-inspired module that integrates delayed sensory feedback (Fig.~\ref{fig:model}). Our results show that (i) multiplexed cerebellar outputs improves the accuracy of online correction, and (ii) routing delayed feedback through the cerebellar loop can facilitate faster and more stable adaptation while also reproducing key neural population patterns observed experimentally.

Together, these findings provide a mechanistic account of how cerebellar processing transforms delayed sensory signals into effective feedback, unifying predictive control and error-driven learning within a single framework.

The implementation used in this work is publicly available at
\url{https://github.com/acarolinafilipe/kinematics_error_cerebellar_feedback}.

\section{Related Work}
The cerebellum has long been proposed to implement forward models for predicting sensory consequences of motor commands, enabling error-driven adaptation~\cite{Marr1969,albus1971theory,Ito1984,bastian2006learning}. Behavioral studies show that motor adaptation operates on multiple timescales, combining rapid error correction with slower internal model updates~\cite{smith2006interacting,krakauer2000learning}. 

At the neural level, motor cortex exhibits structured population dynamics during adaptation, including dual-phase responses associated with feedforward and feedback processes~\cite{perich2018rapid,feulner2025neural}. Recurrent neural networks (RNNs) have been widely used to model such dynamics and generate realistic motor behavior~\cite{sussillo2015neural}. Recent work has incorporated feedback signals into RNNs to enable online correction, often by providing sensory feedback directly to cortical units~\cite{feulner2025neural,kaleb2024feedback}.

Alternatively, some approaches integrate cerebellum-inspired modules that learn predictive mappings and provide corrective signals to cortical networks~\cite{pemberton2024cerebellar,agueci2025distributed,pemberton2021cortico,boven2023cerebro}. However, the functional role and routing of delayed sensory feedback remain unclear. In this work, we explicitly compare direct cortical feedback with cerebellar-mediated feedback, showing that the latter better captures both behavioral adaptation and neural dynamics.

\section{Methods}

\subsection{Task}
We construct a synthetic instructed-delay centre-out reaching task inspired by monkey experiments~\cite{perich2018rapid}. The initial and target positions, $\mathbf{p}_{\text{start}}, \mathbf{p}_{\text{end}} \in \mathbb{R}^2$, are independently sampled from a uniform distribution $\mathcal{U}(-6, 6)$ along each axis. Desired velocity trajectories $[v_x, v_y]$ are generated using a sigmoid temporal profile:

\begin{equation}
f(t) = \frac{1}{1 + \exp(-\kappa t)},
\label{eq:sigmoid}
\end{equation}
with $\kappa = 10\,\mathrm{s}^{-1}$, producing smooth, bell-shaped reaching dynamics.
Each trial lasts 300 time steps ($\approx 3\,\mathrm{s}$) and includes a variable instructed-delay period sampled uniformly from $[1.2, 1.7]\,\mathrm{s}$. The hold signal is encoded as an additional binary input channel using a Heaviside step function active during the delay window. The resulting task has a 3D input (2D position target + hold signal) and a 2D velocity output.

To evaluate online correction and adaptation, we introduce a visuomotor rotation (VR) perturbation of $30^\circ$, following established paradigms in human and monkey studies~\cite{krakauer2000learning,perich2018rapid}.

\subsection{Network Architecture}

We model the cortical component as a continuous-time recurrent neural network (RNN) discretized using Euler integration. Let $h(t)\in\mathbb{R}^N$ denote the cortical state and $x(t)$ the task input. The network dynamics are

\begin{equation}
h(t+1)=h(t)+\frac{dt}{\tau}
\left(
-h(t)+W_h\Phi(h(t))+W_{in}x(t)+W_{fb}C(t)
\right),
\end{equation}
where $\Phi(\cdot)$ is a ReLU nonlinearity and $W_{fb}$ projects feedback (cerebellar) signals back to the cortex. The network output corresponds to hand velocity, leveraging the direct correlation between motor cortex activity and hand velocity~\cite{feulner2025neural}:

\begin{equation}
y(t)=W_{out}\Phi(h(t)).
\end{equation}
Position is obtained by integrating velocity, $p(t)=p(t-1)+dt\,y(t).$

The cerebellar module receives delayed cortical activity and delayed sensory feedback and produces a corrective signal,

\begin{equation}
C(t)=W_{PF}\Phi\!\left(
W_{MF}[h(t-1),s(t-\Delta)]
\right),
\end{equation}
where $[\cdot,\cdot]$ denotes concatenation. The mossy-fibre projection $W_{MF}$ implements a 1:20 expansion~\cite{Marr1969,albus1971theory,Ito1984}. Delayed sensory feedback is provided as
$s(t-\Delta)=\epsilon(t-\Delta),$ with $\epsilon(t)=p^*(t)-p(t),$ where $p^*(t)$ denotes the desired trajectory. Unless otherwise stated, the feedback delay was fixed at $\Delta=120$ ms.

\subsection{Adaptation and Learning}

Cortical and cerebellar learning operate through distinct mechanisms. Cortical adaptation is implemented using a local eligibility-trace rule inspired by biologically plausible feedback-driven plasticity~\cite{feulner2025neural,kaleb2024feedback}. Eligibility traces accumulate recent neural activity,

\begin{equation}
r_i(t)=\sum_{t'<t}\Phi(h_i(t')),
\end{equation}
and recurrent cortical weights are updated according to

\begin{equation}
\Delta W_{ji}
\propto
\sum_t\sum_k
W_{fb,jk}C_k(t)r_i(t).
\end{equation}

Updates are accumulated throughout a trial and applied at its end. This learning rule relies only on locally available activity traces and cerebellar feedback signals.
The loss function of the cortical RNN is defined to minimize the position error $\epsilon$ between the predicted and target trajectories.

Depending on the experimental condition, the target $y^*(t)$ corresponds to velocity, position, position error, or a multiplexed representation.
The cerebellar loss is defined as:

\begin{equation}
L_{cb}
=
\frac{1}{T}
\sum_{t=1}^{T}
\|C(t)-y^*(t-\Delta)\|^2.
\end{equation}
This formulation ensures that cerebellar learning is driven only by biologically available delayed feedback, consistent with climbing-fiber-mediated teaching signals~\cite{pemberton2024cerebellar,Marr1969,albus1971theory}.
Only the parallel-fibre weights $W_{PF}$ are updated using gradient descent, while gradients are computed with respect to the delayed target. Mossy-fibre weights $W_{MF}$ remain fixed.

\subsection{Training Procedure}

We adopt a two-stage training protocol designed to reproduce motor adaptation dynamics~\cite{perich2018rapid,feulner2025neural,pemberton2024cerebellar}. The RNN contained $N=50$ recurrent units. All cortical parameters $(W_{in}$, $W_h$,$W_{out},W_{fb})$ were initialized from a uniform distribution
$U(-1/\sqrt{l},1/\sqrt{l})$, where $l$ denotes the layer fan-in~\cite{he2015delving}. Simulations used a time step of $dt=10$ ms and a time constant of $\tau=50$ ms.

Training proceeded in two stages. During an initial pre-training phase (100 epochs), feedback projections $W_{fb}$ were held fixed, allowing the cerebellar module to learn predictive representations without altering cortical dynamics. In a second phase (500 epochs), cortical adaptation and cerebellar learning were enabled simultaneously, allowing closed-loop cortico-cerebellar interactions to emerge. Cerebellar learning used the Adam optimizer~\cite{kingma2014adam} with learning rate $10^{-3}$ and parameters $(\beta_1,\beta_2)=(0.9,0.999)$. Training was performed with mini-batches of 20 trials. Cortical eligibility traces were accumulated online during each trial, whereas cortical weight updates were applied at the end of the trial. Cerebellar gradient updates are accumulated at each timestep but are supervised by delayed error signals arriving after a fixed delay $\Delta$.

\section{Results}

\subsection{Multiplexing Enables Online Compensation for External Perturbations}

\begin{figure}[h!]
\centering
\includegraphics[width=0.85\textwidth]{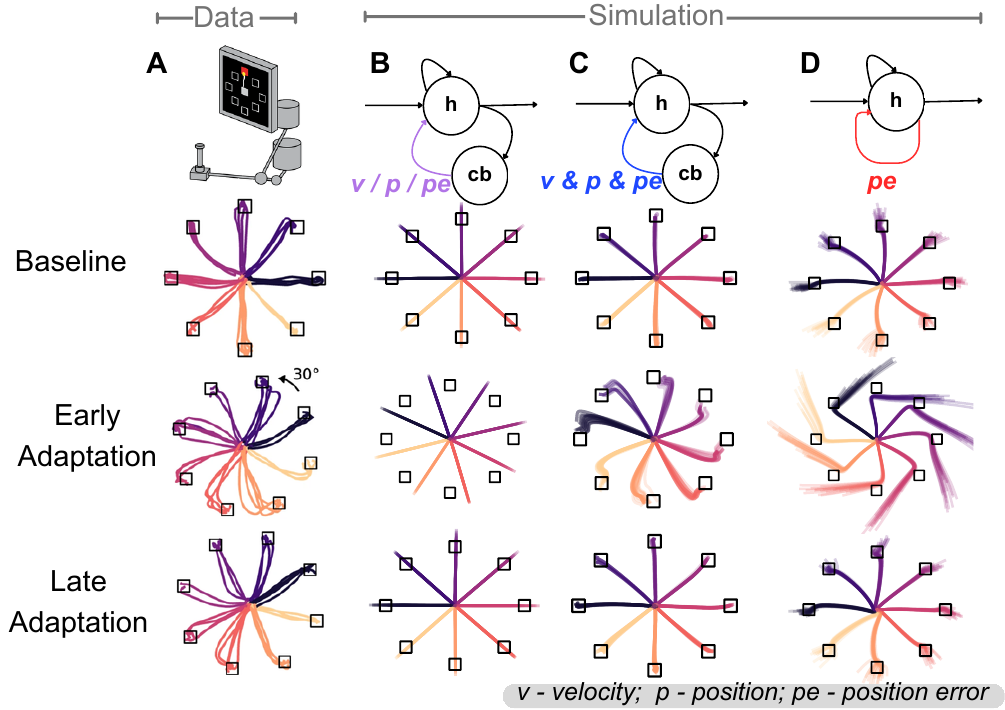}
\caption{\textbf{Multiplexed representations are required for robust online error correction.} 
    \textbf{(A) Experimental reaching behavior:} Hand trajectories across three learning phases: baseline (initial 30 trials), early adaptation (first 10 perturbation trials), and late adaptation (final 30 trials of a 150-trial epoch). Trajectories are color-coded by target direction. Data adapted from \cite{perich2018rapid}. 
    \textbf{(B) Single-variable predictive models:} Model predictions on the test set across all phases. Performance is invariant regardless of whether the cerebellar (cb) module predicts velocity, position, or position error independently; notably, single-variable models fail to exhibit corrective re-aiming. 
    \textbf{(C) Multiplexed model:} When the cerebellum integrates velocity, position, \textit{and} position error, the system successfully reproduces the characteristic re-aiming behavior observed during early biological adaptation. 
    \textbf{(D) Direct feedback baseline:} Performance in the absence of the cerebellar module, where control is driven exclusively by the delayed position error signal, resulting in systematic instability and poor target acquisition.}
    \label{fig:trajectories_comparison}
\end{figure}

We compared our framework against three alternative control architectures during delayed reaching: single-variable predictive models, a direct delayed-feedback controller, and a multiplexed predictive model (Figure \ref{fig:trajectories_comparison}).

Only the multiplexed architecture reproduced the rapid online re-aiming observed in monkey behavior during early adaptation (Figure \ref{fig:trajectories_comparison}C). Experimental trajectories showed curved movements that were corrected online toward the target after perturbation. In contrast, single-variable models predicting either position, velocity, or position error failed to generate corrective redirection (Figure \ref{fig:trajectories_comparison}B). Although trajectories remained smooth, they systematically drifted parallel to the target direction, indicating that prediction of a single kinematic variable is insufficient for robust control under delay.

The multiplexed model combined position, velocity, and position error within the cerebellar module, enabling stable and accurate online corrections. Position encoded the current limb state, velocity provided predictive compensation for sensory delay, and position error generated corrective drive toward the target. Together, these signals supported accurate state estimation and rapid correction during perturbation.

Removing the cerebellar predictive module entirely led to strong instability (Figure \ref{fig:trajectories_comparison}D). Control based only on delayed position error produced repeated overcorrections and poor target acquisition, consistent with classical delayed feedback systems.

Quantitative results confirmed these observations (Table~\ref{tab:model_errors}). Single-variable models showed larger task errors and overshoot during early adaptation, whereas the multiplexed architecture maintained low directional and lateral errors across all phases.

Together, these results indicate that successful online correction under delay cannot be achieved through prediction of a single kinematic variable. Instead, effective control relies on a high-dimensional multiplexed representation combining predictive and error-related signals, consistent with evidence that cerebellar cortex encodes such integrated outputs~\cite{popa2019cerebellum}. Although single-variable models can produce smooth, task-complete movements, only multiplexed representations support the rapid, goal-directed corrections characteristic of biological motor behavior.

\begin{table}[h!]
\centering
\caption{Performance of different cerebellar prediction and sensory feedback architectures across adaptation stages. Results are averaged across 3 seeds. Takeoff error is reported in angular deviation (deg), while task error is decomposed into directional and lateral components relative to the movement trajectory (cm). Values are reported as mean$_{\pm\mathrm{std}}$.}
\label{tab:model_errors}

\begin{tabular}{llcccccc}
\toprule
\multicolumn{2}{c}{Model architecture} &
\multicolumn{3}{c}{Takeoff error (deg)} &
\multicolumn{3}{c}{Task error(along traj., lateral) (cm)} \\
\cmidrule(lr){1-2}
\cmidrule(lr){3-5}
\cmidrule(lr){6-8}

\shortstack{Cerebellum\\prediction} &
\shortstack{Sensory\\feedback} &
Baseline &
\shortstack{Early\\adap.} &
\shortstack{Late\\adap.} &
Baseline &
\shortstack{Early\\adap.} &
\shortstack{Late\\adap.} \\
\midrule

\multicolumn{1}{c}{\textcolor{purple!50}{v}} & \multicolumn{1}{c}{--} &
$0_{\pm1}$ & $28_{\pm4}$ & $0_{\pm2}$ &
$(1_{\pm0.4},\,0_{\pm0.8})$  &
$(1.5_{\pm0.5},\,2.2_{\pm0.8})$&
$(1.1_{\pm1},\,0_{\pm0.5})$ \\

\multicolumn{1}{c}{\textcolor{purple!50}{p}} & \multicolumn{1}{c}{--} &
$0_{\pm1}$ & $30_{\pm5}$ & $0_{\pm2}$  &
$(0.8_{\pm0.6},\,0_{\pm0.5})$  &
$(1_{\pm1},\,2.2_{\pm0.8})$&
$(0.8_{\pm1},\,0_{\pm0.5})$ \\

\multicolumn{1}{c}{\textcolor{purple!50}{pe}} & \multicolumn{1}{c}{--} &
$0_{\pm0.5}$ & $30_{\pm3}$ & $0_{\pm0.5}$ &
$(0_{\pm0.3},\,0_{\pm0.1})$  &
$(0_{\pm0.3},\,1.5_{\pm0.9})$&
$(0_{\pm0.3},\,0_{\pm0.3})$ \\

\multicolumn{1}{c}{\textcolor{blue}{v + p + pe}} & \multicolumn{1}{c}{--} &
$0_{\pm0.5}$ & $29_{\pm2}$ & $0_{\pm0.5}$ &
$(0_{\pm0.2},\,0_{\pm0.1})$ &
$(0.1_{\pm0.2},\,0.3_{\pm0.2})$ &
$(0_{\pm0.1},\,0_{\pm0.1})$\\

\multicolumn{1}{c}{--} & \multicolumn{1}{c}{\textcolor{red}{pe}} &
$0_{\pm2}$ & $30_{\pm5}$ & $3_{\pm2}$ &
$(1.8_{\pm0.5},\,0_{\pm0.5})$  &
$(2.7_{\pm1},\,3.5_{\pm1.8})$&
$(1.5_{\pm0.6},\,0_{\pm0.6})$ \\

\multicolumn{1}{c}{v + p + pe} &
\multicolumn{1}{c}{\textcolor{green!50!black}{pe ($\rightarrow$ h)}} &
$0_{\pm0.5}$ & $29_{\pm2}$ & $0_{\pm0.5}$ &
$(0_{\pm0.1},\,0_{\pm0.1})$ &
$(0.1_{\pm0.6},\,1.8_{\pm1.5})$ &
$(0_{\pm0.1},\,0_{\pm0.1})$\\

\hline
\hline

\multicolumn{1}{c}{v + p + pe} &
\multicolumn{1}{c}{\textcolor{orange}{pe($\rightarrow$ cb)} } &
$0_{\pm0.5}$ & $30_{\pm1}$ & $0_{\pm0.5}$ &
$(0_{\pm0.1},\,0_{\pm0.1})$ &
$(0.1_{\pm0.2},\,0.2_{\pm0.2})$ &
$(0_{\pm0.1},\,0_{\pm0.1})$\\
\bottomrule
\end{tabular}
\end{table}

We therefore focus subsequent analyses on the multiplexed architecture for studying sensory feedback integration under delay.

\subsection{Error-Based Feedback Enables Rapid Trial-by-Trial Adaptation}

A hallmark of biological motor adaptation is the ability to rapidly reduce task error across trials while more gradually updating internal models~\cite{taylor2012role,bastian2006learning}. Empirically, this process is characterized by a dissociation between fast endpoint error correction and slower changes in initial movement direction (take-off angle), the latter reflecting internal model adaptation often attributed to cerebellar processing~\cite{perich2018rapid}.

\begin{figure}[h!]
\centering
\includegraphics[width=0.99\textwidth]{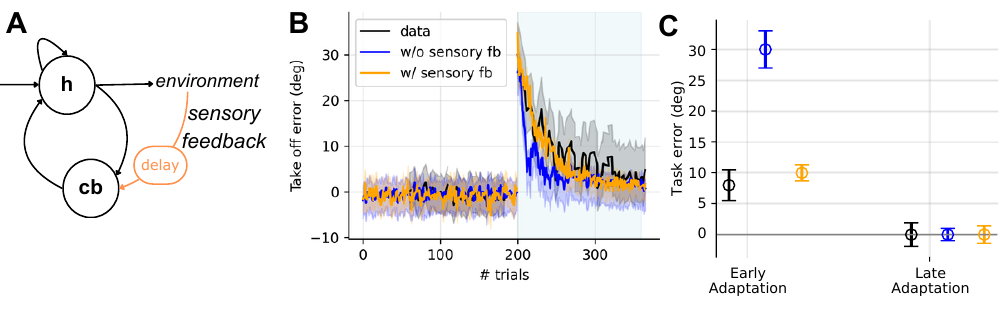}
\caption{\textbf{Error-driven adaptation dynamics.} 
(A) Model architecture with delayed sensory feedback to the cerebellum. 
(B) Take-off angle error across baseline (white) and adaptation (blue) phases. 
(C) Task error (endpoint angular error) during early (first 10 trials) and late (last 30 trials) adaptation, comparing models and experimental data.}
\label{fig:delay}
\end{figure}

To investigate the mechanisms underlying this behavior, we compared two model variants: (i) the proposed architecture with delayed sensory feedback to the cerebellar module (Fig.~\ref{fig:delay}A), and (ii) a control model without this feedback pathway. Both models were evaluated against experimental data~\cite{perich2018rapid}.

Figure~\ref{fig:delay}B shows the evolution of the take-off error. The model with delayed sensory feedback closely matches the gradual reduction observed in the experimental data. In contrast, the model without cerebellar sensory input adapts substantially faster, exhibiting a steeper decline in take-off error. This difference arises because, in the absence of delayed sensory feedback, learning is driven solely by optimization pressure on cortical dynamics, leading to rapid but less biologically realistic adjustments. 

By comparison, the proposed model receives delayed sensory signals corresponding to task-relevant error information, consistent with known feedback delays in biological systems~\cite{honda2012adaptation,botzer2013feedback}. This allows ongoing movements to be corrected online while reducing the need for immediate changes in the feedforward command, resulting in a slower and more realistic adaptation of initial movement direction.

Task error dynamics further support this interpretation (Fig.~\ref{fig:delay}C). Both models achieve near-zero error at late adaptation, indicating that they successfully learn the perturbation. However, during early adaptation, the model with delayed sensory feedback exhibits significantly lower endpoint error, closely matching experimental observations. This suggests that cerebellar feedback enables rapid online correction of movements before full internal model adaptation occurs, consistent with theories of feedback-driven motor learning~\cite{feulner2025neural}.

Overall, these results demonstrate that incorporating delayed sensory feedback into the cerebellar pathway is critical to reproducing the characteristic dual-timescale adaptation observed in biological motor behavior: fast reduction of task error alongside slower recalibration of feedforward motor commands.

\subsection{Sensory Feedback Shapes Dual-Phase Cortical Dynamics}

Next, we examine whether sensory feedback signals should be provided directly to the cortical RNN, as proposed in previous modeling studies~\cite{feulner2025neural,kaleb2024feedback}, or routed exclusively through the cerebellar module as in our framework. In particular, we asked how these alternative pathways influence cortical population dynamics during adaptation.

Experimental recordings from primary motor cortex (M1) show that adaptation is accompanied by a characteristic dual-phase response: an early peak associated with feedforward control, followed by a later peak reflecting feedback-driven corrections~\cite{perich2018rapid}. These two temporally dissociable components are thought to reflect the interaction between predictive feedback and delayed sensory feedback.

To test whether our model reproduces this phenomenon, we compared three conditions: (i) no sensory feedback, (ii) sensory feedback routed through the cerebellar module (proposed model), and (iii) sensory feedback provided directly to the RNN in parallel with cerebellar feedback.

\begin{figure}[h!]
\centering
\includegraphics[width=0.96\textwidth]{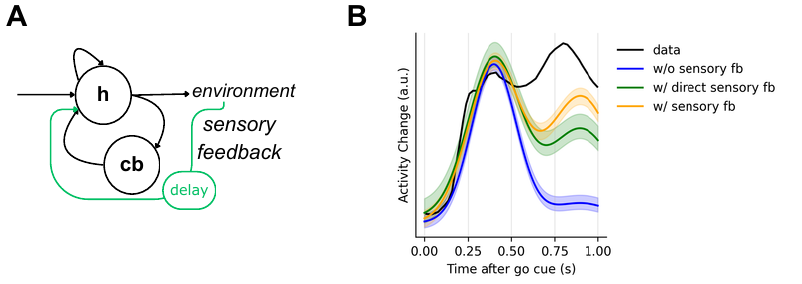}
\caption{\textbf{Cortical dynamics under different feedback architectures.}
(A) Model variant with both cerebellar feedback and direct delayed sensory input to the RNN.
(B) Average change in RNN activity across time, computed from population activity $r(t)$ obtained by averaging hidden-unit activity across neurons and trials;
and comparison with experimental recordings~\cite{perich2018rapid}. Neural population activity is obtained by averaging firing rates across neurons, and both model and data are aligned to movement onset (Go cue) for temporal comparison.}
\label{fig:waves}
\end{figure}

Figure~\ref{fig:waves}B shows that models incorporating sensory feedback (conditions ii and iii) reproduce the experimentally observed late feedback-related peak in activity. In contrast, the model without sensory feedback exhibits only a single early peak, corresponding to feedforward dynamics.

Importantly, routing sensory feedback through the cerebellar module is sufficient to recover the dual-phase structure, indicating that explicit direct feedback to the cortical RNN is not required. This suggests that cerebellar processing can mediate the transformation of delayed sensory signals into effective feedback for cortical dynamics.

Overall, these results support a model in which the cerebellum plays a central role in shaping feedback-related cortical activity, enabling the emergence of biologically consistent dual-phase dynamics during motor adaptation.

\subsection{Error-Dependent Modulation of Feedback Dynamics}

Finally, we compared network dynamics across baseline and adaptation to assess how feedback-related activity depends on task error. Experimental evidence suggests that feedback-driven neural responses are prominent only when errors are present, whereas behavior in the absence of perturbations is largely governed by feedforward control~\cite{perich2018rapid}.

\begin{figure}[h!]
\centering
\includegraphics[width=0.9\textwidth]{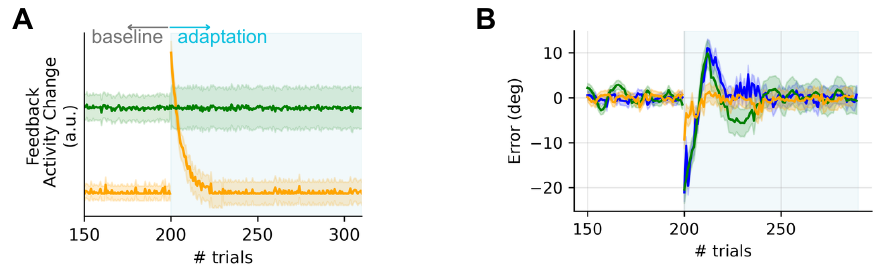}
\caption{\textbf{Error-dependent feedback dynamics and adaptation performance.}
(A) Feedback-related activity change measured 800 ms after the go cue across baseline and adaptation phases.
(B) Task error across trials for models without sensory feedback (blue), with cerebellar feedback (orange), and with direct sensory feedback to the RNN (green).}
\label{fig:errors_fb}
\end{figure}

In Fig.~\ref{fig:errors_fb}A, we quantify the time course of feedback-related activity within a fixed window (800 ms after the go cue) for two model variants: (i) sensory feedback routed through the cerebellum (proposed model) (orange), and (ii) direct sensory feedback to the RNN (green). 

The direct-feedback model exhibits sustained feedback-related activity across all phases, including baseline, early adaptation, and late adaptation. In contrast, the cerebellar-feedback model shows a transient increase in feedback activity specifically during early adaptation, when task errors are largest, followed by a gradual decay as performance improves. This behavior is consistent with an error-gated feedback mechanism, where feedback contributions diminish once accurate feedforward control is established.

Figure~\ref{fig:errors_fb}B compares the task error between trials for the different variants of the model. The cerebellar-feedback model exhibits a smooth and stable reduction in error, rapidly converging to near-zero levels. In contrast, the model without delayed sensory feedback adapts more slowly, requiring substantially more trials to achieve similar performance. The model with direct sensory feedback shows less stable learning dynamics, with increased variability during adaptation.

Together, these findings indicate that a cerebellar-mediated feedback loop is essential for both biological consistency and algorithmic stability. Rather than letting raw sensory data directly disrupt core network dynamics, this modular processing regulates the timing and magnitude of error signals, providing a robust architectural blueprint for stabilizing adaptive learning under feedback delays.

\section{Conclusion}

In this work, we investigated the role of delayed sensory feedback in shaping motor adaptation within a cortico-cerebellar recurrent neural network model. Our results show that routing sensory feedback through a cerebellar module enables the emergence of key features of biological motor learning, including rapid trial-by-trial error reduction, slower adaptation of movement initiation, and dual-phase cortical dynamics.

In contrast, models lacking delayed feedback or receiving direct sensory input at the cortical level exhibit less biologically consistent behavior, such as overly rapid adaptation, persistent feedback activity, or unstable learning dynamics. By incorporating delayed error signals, the proposed model captures an error-dependent gating of feedback, consistent with theories of cerebellar forward models and prediction error minimization~\cite{bastian2006learning,izawa2011learning,popa2019cerebellum}.

Overall, our findings support the view that the cerebellum plays a central role in transforming delayed sensory information into effective feedback for motor control, supporting learning while preserving flexibility. From an artificial intelligence and control systems perspective, this architecture addresses a fundamental challenge in autonomous agents: managing system latency. In robotics and deep reinforcement learning, feedback delays can lead to overcorrections or degraded performance. Our results suggest that a modular division of labor, i.e., separating instantaneous, predictive error correction from a core feedforward controller, may provide a robust, bio-inspired blueprint for designing adaptive artificial systems operating under real-world latency constraints.

\begin{credits}
\subsubsection{\ackname}
A.C.F.’s work was supported by the ”la Caixa” Foundation through a Doctoral INPhINIT Retaining Fellowship (Grant CCA 0404020203). This work was partially supported by NOVA LINCS UID/04516/2025 (FCT.IP), UID/PRR/04516/2025, UID/PRR2/04516/2025 (NextGenerationEU),
Neuraspace AI Fights Space Debris project (project code C626449889-00463050, operation code 2022-C05i0101-02), co-funded by Recovery and Resilience Plan and NextGeneration EU Funds, www.recuperarportugal.gov.pt.

\subsubsection{\discintname}
The authors have no competing interests to declare that are
relevant to the content of this article.
\end{credits}

%
%
%
\bibliographystyle{splncs04}
\bibliography{references}

\end{document}